\documentclass[journal]{IEEEtran}
\usepackage{cite}

\usepackage{graphicx}
\usepackage{tabularx}
\usepackage{multirow}
\ifCLASSINFOpdf
\else
\fi
\usepackage{amsmath}
\usepackage{amsfonts}
\usepackage{algorithm}
\usepackage{algorithmic}

\usepackage{array}

\newcolumntype{Y}{>{\centering\arraybackslash}X}

\begin{document}
%
\title{Query-adaptive Image Retrieval by \\Deep Weighted Hashing}
%
%

\author{Jian Zhang
		and Yuxin Peng
\thanks{This work was supported by National Natural Science Foundation of China under Grants 61371128 and 61532005.}
\thanks{The authors are with the Institute of Computer Science and Technology, Peking University, Beijing 100871, China. Corresponding author: Yuxin Peng (e-mail: pengyuxin@pku.edu.cn).}
}

\maketitle

\begin{abstract}
Hashing methods have attracted much attention for large scale image retrieval. Some deep hashing methods have achieved promising results by taking advantage of the strong representation power of deep networks recently. However, existing deep hashing methods treat all hash bits equally. On one hand, a large number of images share the same distance to a query image due to the discrete Hamming distance, which raises a critical issue of image retrieval where fine-grained rankings are very important. On the other hand, different hash bits actually contribute to the image retrieval differently, and treating them equally greatly affects the retrieval accuracy of image. To address the above two problems, we propose the query-adaptive deep weighted hashing (QaDWH) approach, which can perform fine-grained ranking for different queries by weighted Hamming distance. First, a novel deep hashing network is proposed to learn the hash codes and corresponding class-wise weights jointly, so that the learned weights can reflect the importance of different hash bits for different image classes. Second, a query-adaptive image retrieval method is proposed, which rapidly generates hash bit weights for different query images by fusing its semantic probability and the learned class-wise weights. Fine-grained image retrieval is then performed by the weighted Hamming distance, which can provide more accurate ranking than the traditional Hamming distance. Experiments on four widely used datasets show that the proposed approach outperforms eight state-of-the-art hashing methods.
\end{abstract}

\begin{IEEEkeywords}
Deep weighted hashing, query-adaptive, image retrieval.
\end{IEEEkeywords}

%
\IEEEpeerreviewmaketitle

\section{Introduction}
\IEEEPARstart{W}ITH rapid growth of multimedia data on web, retrieving the relevant multimedia content from a massive database has been an urgent need, yet still remains a big challenge. Hashing methods map multimedia data into short binary codes to utilize the storage and computing efficiency of Hamming codes, thus they have been receiving increasing attentions in many multimedia application scenarios, such as image retrieval~\cite{gionis1999similarity,7298947,wang2010sequential,weiss2009spectral,chen2017nonlinear,li2013spectral,kafai2014discrete,zhang2014prior}, video retrieval~\cite{liong2016deep,hao2017stochastic} and cross media retrieval~\cite{ding2017cross,wang2015learning}. Generally speaking, hashing methods aim to learn mapping functions for multimedia data, so that similar data are mapped to similar binary codes. These hashing methods can be divided into two categories, namely unsupervised methods and supervised methods.

Unsupervised methods are proposed to design the hash functions without using image labels, and they can be divided into data independent methods and data dependent methods. The representative data independent method is Locality Sensitive Hashing (LSH)~\cite{gionis1999similarity}, which maps data into the binary codes by random linear projections. There are several extensions of LSH, such as SIKH~\cite{raginsky2009locality} and Multi-probe LSH~\cite{lv2007multi}. Data dependent methods try to learn hash functions by analyzing the data properties, such as manifold structures and data distributions. For example, Spectral Hashing (SH)~\cite{weiss2009spectral} is proposed to design the hash codes to be balanced and uncorrelated. Anchor Graph Hashing (AGH)~\cite{liu2011hashing} is proposed to use anchor graphs to discover neighborhood structures. Gong et al. propose Iterative Quantization (ITQ)~\cite{gong2011iterative} to learn hash functions by minimizing the quantization error of mapping data to the vertices of a binary hypercube. Topology Preserving Hashing (TPH)~\cite{zhang2013topology} is proposed to preserve the consistent neighborhood rankings of data points in Hamming space. Irie et al. propose Locally Linear Hashing (LLH)~\cite{irie2014locally} to utilize locality-sensitive sparse coding to capture the local linear structures and then recover these structures in Hamming space.

\begin{figure}[tb]
	\centering
	\includegraphics[width=0.5\textwidth]{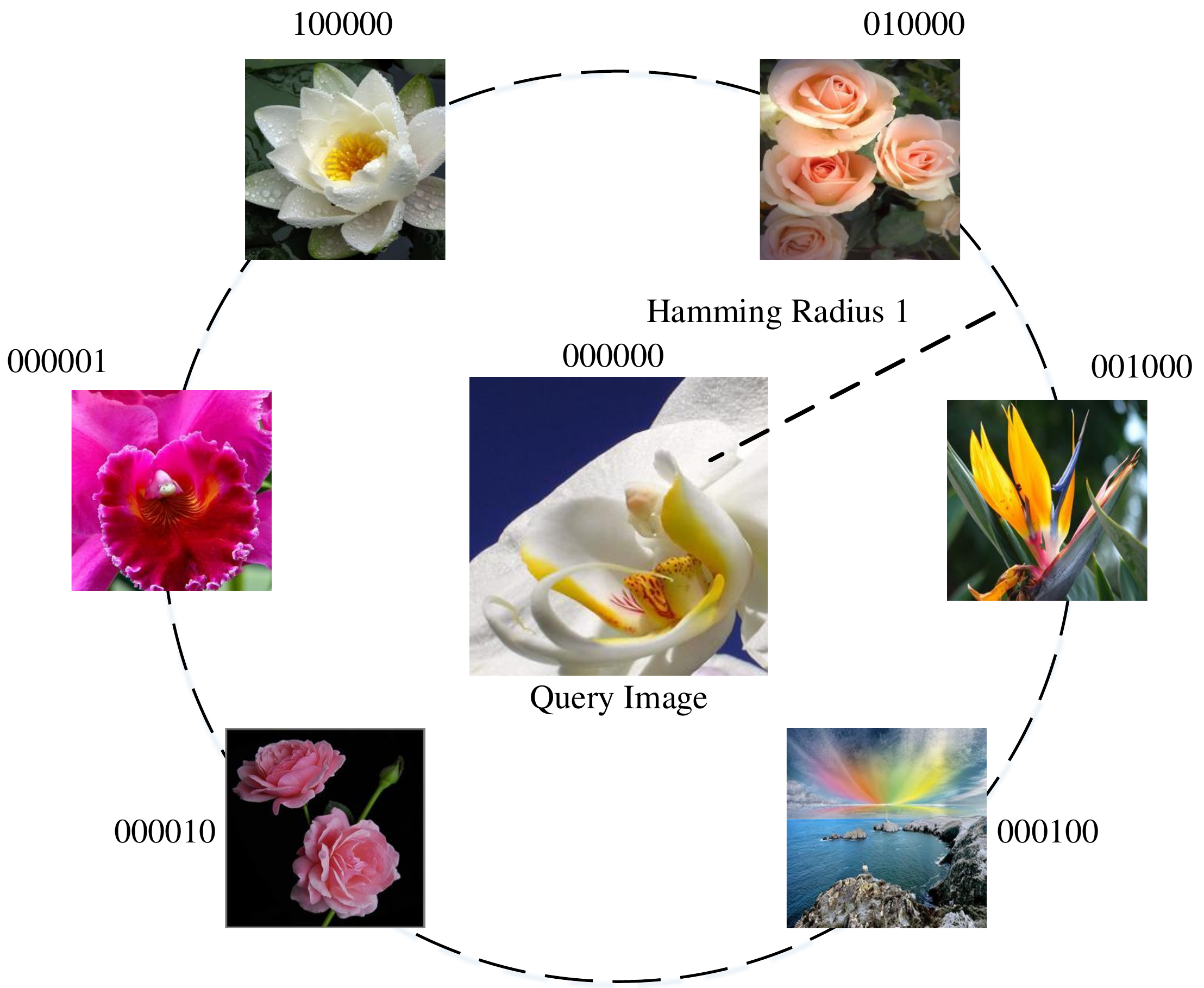}
	\caption{An illustration of the discrete Hamming distance. Here we use an example of 6bits hash codes, suppose the query image has a hash code of $000000$, there are 6 images within Hamming radius 1 with query image, however they differ in different bits, traditional Hamming distance cannot perform fine-grained ranking among them.}
	\label{intro}
\end{figure}

Unsupervised methods can search nearest neighbors under a certain kind of distance metric (e.g. $L_2$ distance). However, the neighbors in feature space may not be semantically similar. Therefore, supervised hashing methods are proposed, which leverage the semantic information to generate effective hash codes. Binary Reconstruction Embedding (BRE)~\cite{kulis2009learning} is proposed to learn the hash functions by minimizing the reconstruction error between the original distances and the reconstructed distances in the Hamming space. Wang et al. propose Semi-supervised Hashing (SSH)~\cite{wang2010sequential} to learn hash functions by minimizing the empirical error over the labeled data while maximizing the information entropy of generated hash codes over both labeled and unlabeled data. Liu et al. propose Supervised Hashing with Kernels (KSH)~\cite{liu2012supervised} to learn the hash codes by preserving the pairwise relationship between data samples provided by labels. Order Preserving Hashing (OPH)~\cite{wang2013order} and Ranking Preserving Hashing (RPH)~\cite{wang2015ranking} are proposed to learn hash functions by preserving the ranking information, which is obtained based on the number of shared semantic labels between images. Supervised Discrete Hashing (SDH)~\cite{7298598} is proposed to leverage label information to obtain hash codes by integrating hash code generation and classifier training.

Although aforementioned unsupervised and supervised methods have achieved considerable progress, the image representations of these methods are hand-crafted features (e.g. GIST~\cite{oliva2001modeling}, Bag-of-Visual-Words~\cite{fei2005bayesian}), which can not well represent the images' semantic information. Inspired by the successful applications of deep networks on image classification and object detection~\cite{krizhevsky2012imagenet}, some deep hashing methods have been proposed recently to take advantage of the superior feature representation power of deep networks. Convolutional Neural Network Hashing (CNNH)~\cite{xia2014supervised} is a two-stage framework, which is designed to learn the fixed hash codes in the first stage by preserving the pairwise semantic similarity, and learn the hash functions based on the learned hash codes in the second stage. Although the learned hash codes can guide the feature learning, the learned features cannot give feedback for learning better hash codes. To overcome the shortcomings of the two-stage learning scheme, some approaches have been proposed to perform simultaneously image feature and hash code learning. Lai et al. propose Network In Network Hashing (NINH)~\cite{7298947} to use a triplet ranking loss~\cite{schultz2003learning} to capture the relative similarities of images. NINH is a one-stage supervised method, thus the image representation learning and hash code learning can benefit each other in the deep architecture. Some similar ranking-based deep hashing methods~\cite{yaodeep, zhang2015bit,zhao2015deep} have been proposed recently, which are also designed to preserve the ranking information obtained by labels. Besides the triplet ranking based methods, some pairwise based deep hashing methods~\cite{li2015feature, zhu2016deep} are also proposed, which try to preserve the semantic similarities provided by pairwise labels.

Although deep hashing methods have achieved promising results on image retrieval, existing deep hashing methods~\cite{xia2014supervised,7298947,zhang2015bit} treat all hash bits equally. However, Hamming distances are discrete integers, so there are often a large number of images sharing the equal Hamming distances to a query, which raises a critical issue of image retrieval where fine-grained rankings are very important. An example is illustrated in Figure~\ref{intro}, given a query image with hash code $000000$, there can be 6 images with different hash codes that within Hamming radius 1 of query image, while they differ in different hash bits. Existing deep hashing methods cannot perform fine-grained ranking among them. However, if we know that which bit of hash codes is more important for query image, then we can return a better rankling list.

There exist several traditional hashing methods~\cite{ji2014query, jiang2011lost,jiang2013query,liu2016query,zhang2013binary,zhang2012qsrank} that learn weights for hash codes. However, these methods adopt two-stage frameworks, which first generate hash codes by other methods (e.g. LSH, ITQ), then learn hash weights by analyzing the fixed hash codes and image features. The two-stage scheme causes that the learned weights can't give feedback for learning better hash codes, which limits the retrieval accuracy. Thus we propose the query-adaptive deep weighted hashing (QaDWH) method, which can not only learn hash codes and corresponding class-wise weights jointly, but also perform effective yet efficient query-adaptive fine-grained retrieval. To the best of our knowledge, this is the first deep hashing method that can perform query-adaptive fine-grained ranking. The main contributions of this paper can be concluded as follows:
\begin{itemize}
	\item A novel deep hashing network is designed to learn hash functions and corresponding weights jointly. In the proposed deep network, a hash layer and a class-wise weight layer are designed, of which the hash layer generates hash codes, while the class-wise weight layer learns the class-wise weights for different hash bits. On the top of the hash stream, a weighted triplet ranking loss is proposed to preserve the similarity constraints. With the trained deep network, we can not only generate the binary hash codes, but also weigh the importance of each bit for different image class.
	\item A query-adaptive retrieval method is proposed to perform fine-grained retrieval. Query image's bit-wise hash weights are first rapidly generated by fusing the learned class-wise weights and the predicted query class probability, so that generated weights can reflect different query's semantic property. Based on the weights and generated hash codes, weighted Hamming distance measurement is employed to perform fine-grained rankings for different query images.
\end{itemize}

Extensive experiments on four datasets show that the proposed approach achieves the best retrieval accuracy comparing to eight state-of-the-art hashing methods. The rest of this paper is organized as follows. Section II briefly reviews the related work, section III presents the proposed deep weighted hashing method, section IV shows the experiments on four widely used image datasets, and section V concludes this paper.

\section{Related work}
In this section, we briefly review related work, including some deep hashing methods proposed recently, and traditional weighted hashing methods.
\subsection{Deep Hashing Methods}
 Convolutional Neural Network Hashing (CNNH)~\cite{xia2014supervised} is the first deep hashing method based on convolutional neural networks (cnns). CNNH is composed of two stages: a hash code learning stage and a hash function learning stage. Given a training image set $\mathcal{I} = \{I_1,I_2,\ldots,I_n\}$, in the hash code learning stage (Stage 1), CNNH learns approximate hash codes for training images by optimizing the following loss function:
\begin{equation}
\label{cnnh_obj_1}
\min \limits_H {\|S-\frac{1}{q}HH^T\|^2_F}
\end{equation}
where $\|\cdot\|_F$ denotes the Frobenius norm; $S$ denotes the semantic similarity of image pairs in $\mathcal{I}$, in which $S_{ij}=1$ when image $I_i$ and $I_j$ are semantically similar, otherwise $S_{ij}=-1$; $H \in \{-1,1\}^{n \times q}$ denotes the approximate hash codes. $H$ encodes the approximate hash codes for training images which preserve the pairwise similarities in $S$. Equation~(\ref{cnnh_obj_1}) is difficult to directly optimize, thus CNNH firstly relaxes the integer constraints on $H$ and randomly initializes $H \in [-1,1]^{n \times q}$, then optimizes equation~(\ref{cnnh_obj_1}) using a coordinate descent algorithm, which sequentially or randomly chooses one entry in $H$ to update while keeping other entries fixed. Thus it is equivalent to optimizing the following equation:
\begin{equation}
\min \limits_{H_{\cdot j}}{\|H_{\cdot j}H_{\cdot j}^T-(qS-\sum \limits _{c \not = j}{H_{\cdot c}H_{\cdot c}^T})\|^2_F}
\end{equation}
where $H_{\cdot j}$ and $H_{\cdot c}$ denote the $j$-th and the $c$-th column of $H$ respectively. In the hash function learning stage (Stage 2), CNNH uses deep networks to learn image features and hash functions. Specifically, CNNH adopts the deep framework in~\cite{hinton2012improving} as its basic network, and designs an output layer with sigmoid activation to generate $q$-bit hash codes. CNNH trains the designed deep network in a supervised way, in which the approximate hash codes learned in Stage 1 are used as the ground-truth. However, CNNH is a two-stage framework, where the learned deep features in Stage 2 cannot help to optimize the approximate hash code learning in Stage 1, which limits the performance of hash learning.

Different from the two-stage framework in CNNH~\cite{xia2014supervised}, Network in Network Hashing (NINH)~\cite{7298947} performs image representation learning and hash code learning jointly. NINH constructs deep framework based on the Network in Network architecture~\cite{lin2013network}, with a shared sub-network composed of several stacked convolutional layers to extract image features, as well as a divide-and-encode module encouraged by sigmoid activation function and a piece-wise threshold function to output binary hash codes. During the learning process, instead of generating approximate hash codes in advance, NINH utilizes a triplet ranking loss function to exploit the relative similarity of training images to directly guide hash learning:
\begin{equation}
\label{ninh_loss}
\begin{split}
& l_{triplet}(I,I^+,I^-) = \\
& \max(0,1-(\|b(I)-b(I^-)\|_{\mathcal{H}}-\|b(I)-b(I^+)\|_{\mathcal{H}})) \\
& s.t. \quad b(I), b(I^+), b(I^-) \in \{-1,1\}^q
\end{split}
\end{equation}
where $I,I^+$ and $I^-$ specify the triplet constraint that image $I$ is more similar to image $I^+$ than to image $I^-$ based on image labels; $b(\cdot)$ denotes binary hash code, and $\|\cdot\|_\mathcal{H}$ denotes the Hamming distance. For easy optimization of equation~(\ref{ninh_loss}), NINH applies two relaxation tricks: relaxing the integer constraint on binary hash code and replacing Hamming distance with Euclidean distance. 

There are several extensions based on NINH, such as Bit-scalable Deep Hashing method~\cite{zhang2015bit} further manipulates hash code length by weighing each bit of hash codes. Deep Hashing Network (DHN)~\cite{zhu2016deep} additionally minimizes the quantization errors besides triplet ranking loss to improve retrieval precision. Deep semantic preserving and ranking-based hashing (DSRH)~\cite{yaodeep} introduces orthogonal constraints into triplet ranking loss to make hash bits independent. The above deep hashing methods treat all hash bits equally, which leads to a coarse ranking among images with the same hamming distance and achieves limited retrieval accuracy.

\subsection{Traditional Weighted Hashing Methods}
Hamming distances are discrete integers that can not perform fine-grained ranking for those images sharing same distances with a query image. Some hash bits weighting methods~\cite{ji2014query, jiang2011lost,jiang2013query,liu2016query,zhang2013binary,zhang2012qsrank} are proposed to address this issue. QaRank~\cite{jiang2011lost,jiang2013query} first learns class-specific weights by minimizing the intra-class similarity and maintaining the inter-class relations, then generates query-adaptive weights by using top \textit{k} similar images' labels. QsRank~\cite{zhang2012qsrank} designs a ranking algorithm for PCA-based hashing method, which uses the probability that $\epsilon$-neighbors of query $q$ map to hash code $h$ to measure the ranking score of hash codes $h$. WhRank~\cite{zhang2013binary} proposes a weighted Hamming distance ranking algorithm by data-adaptive weight and query-sensitive bitwise weight. QRank~\cite{ji2014query,liu2016query} learns query-adaptive weights by exploiting both the discriminative power of each hash function and their complement for nearest neighbor search. The aforementioned traditional weighted hashing methods are all two-stage schemes, which take hash codes generated by other methods (such as LSH, SH and ITQ) as input, then learn the weights by analyzing the fixed hash codes and image features. The two-stage scheme causes that the learned weights can't give feedback for learning better hash codes, which limits the retrieval accuracy.

\section{Query-adaptive deep weighted hashing}
\begin{figure*}[ht]
	\centering
	\includegraphics[width=\textwidth]{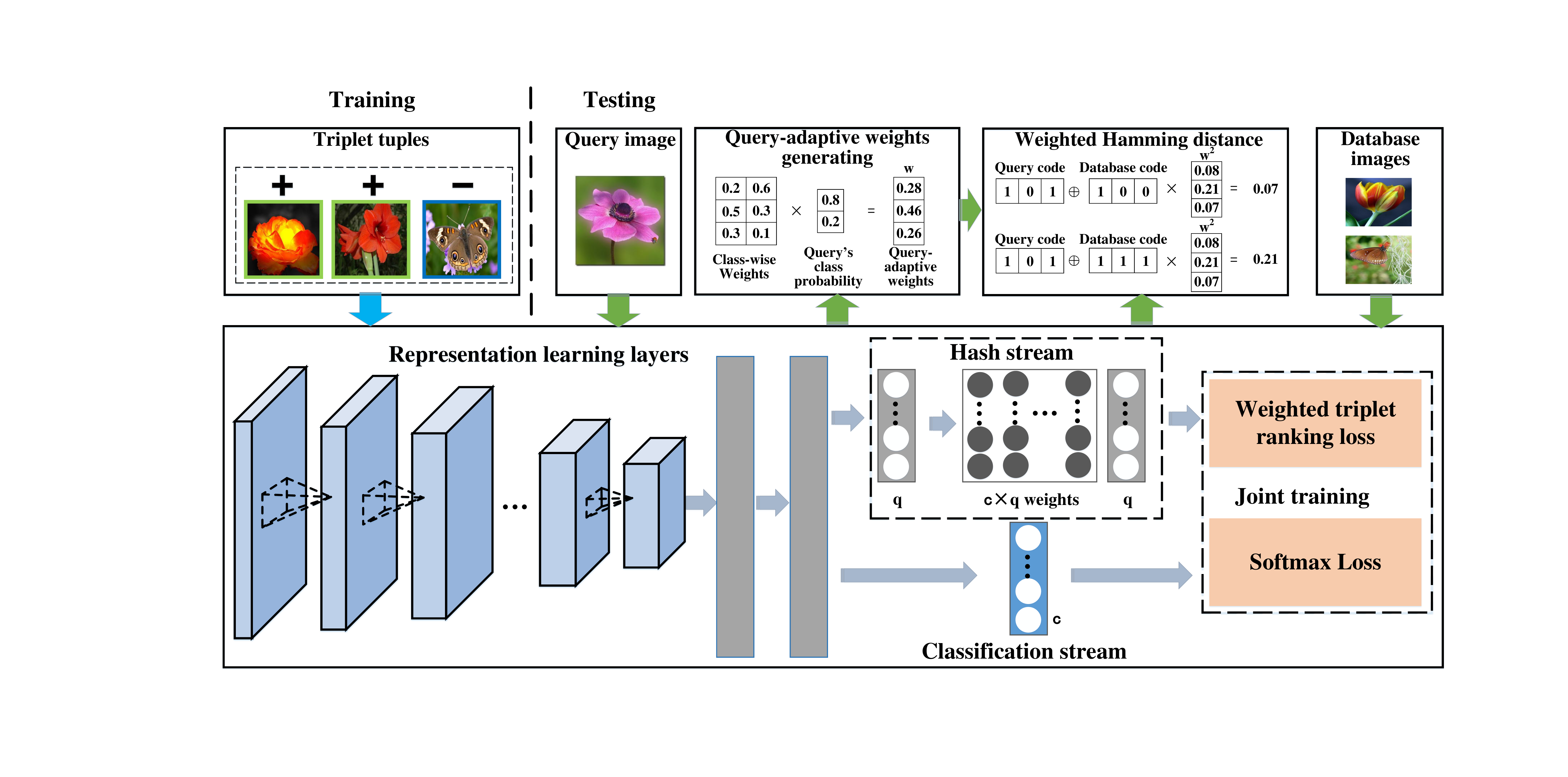}
	\caption{Overview of the proposed deep network architecture, which consists of three main components: the representation learning layers, the hash stream and the classification stream. On the training stage, the images are composed in a triplet form based on the semantic labels. An example of query-adaptive image retrieval is presented on the top right panel.}
	\label{framework}
\end{figure*}

Given a set of $n$ images $\mathcal{X} \in \mathbb{R}^D$. The goal of hashing methods is to learn a mapping function $\mathcal{H}:X \rightarrow \{-1,1\}^q$, which encodes an image $X \in \mathcal{X}$ into a \textit{q}-dimensional binary code $\mathcal{H}(X)$ in the Hamming space, while preserving the semantic similarity of images. In this section, we will introduce the proposed query-adaptive deep weighted hashing (QaDWH) approach. The overall framework is shown in Figure~\ref{framework}, the proposed deep hashing network consists of two streams, namely the hash stream and the classification stream. In the training stage, the hash stream learns the hash functions and the associated weights, while the classification stream preserves the semantic information. In the query stage, the trained network generates compact hash codes and the bit-wise weights for the newly input query images, and then the fine-grained ranking can be performed by the weighted Hamming distance measurement efficiently. In the following of this section, we'll first introduce the proposed deep hashing network and training algorithm, then we'll demonstrate the query-adaptive image retrieval method.
\subsection{Deep Weighted Hashing Network}
As shown in Figure~\ref{framework}, the proposed deep network is composed of the representation learning layers, the hash stream and the classification stream. The representation learning layers serve as a feature extractor, which is a deep network composed of several convolutional layers and fully connected layers. We adopt the VGG-19 network~\cite{Simonyan14c} as the representation learning layers, in which the first 18 layers follow exactly the same settings in the VGG-19 network. The hash stream and the classification stream are both connected with the representation learning layers.

\subsubsection{The Hash Stream}
The hash stream is composed of two layers, the hash code learning layer and the class-wise weight layer. The hash code learning layer is a fully connected layer with \textit{q} neural nodes, its outputs are hash codes defined as:
\begin{equation}
	h(x) = sigmoid(W_h^Tf(x)+v)
\end{equation}
where $f(x)$ is the deep features extracted from the representation learning layers, $W_h$ and $v$ are the parameters in the hash code learning layer. Through the hash code learning layer, the image features $f(x)$ are mapped into $[0,1]^q$. Since the hash codes $h(x) \in [0,1]^q$ are continuous real values, we apply a thresholding function to obtain binary codes:
\begin{equation}
	\label{binarycode}
	b_k(x) = sgn(h_k(x)-0.5), \quad k=1,\cdots,q
\end{equation}

In order to learn the class-wise hash weights, we design a class-wise weight layer connected with the hash code learning layer. The class-wise weight layer is an element-wise layer, which is also associated with image classes. Suppose the number of image classes is \textit{c} and the hash code length is \textit{q}, then the class-wise weight layer is defined as an element-wise layer with $c\times q$ parameters $W=\{w_{ij}| i = 1,\cdots,c, j = 1,\cdots,q\}$. And the output of class-wise weight layer is defined as:
\begin{equation}
	w(h_x) = W(C_x,:)\cdot h_x,\; s.t.\;W \ge 0
\end{equation}
where $h_x$ is the output hash codes of $x$, $C_x$ is the image class index of $x$, and $\cdot$ denotes the element wise product. Here we constraint the weights to be nonnegative. For training images with multiple class, we use average fusion of corresponding weights to perform element wise product. Through the class-wise weight layer, the hash codes of each image are multiplied by its corresponding weights associated with image class.

On the top of the class-wise weight layer, we propose a weighted triplet ranking loss to train the hash stream. For the training images $(\mathcal{X},\mathcal{Y})$, where $\mathcal{Y}$ are the corresponding image labels. We sample a set of triplet tuples depending on the labels, $\mathcal{T}=\{(x_i^L,x_i^{L^+},x_i^{L^-})\}^t_{i=1}$, in which $x_i^L$ and $x_i^{L^+}$ are two similar images with the same labels, while $x_i^L$ and $x_i^{L^-}$ are two dissimilar images with different labels, $t$ is the number of sampled triplet tuples. For the triplet tuple $(x_i^L,x_i^{L^+},x_i^{L^-}), i=1 \cdots t$, the weighted triplet ranking loss is defined as:
\begin{equation}
	\label{rankingterm1}
	\begin{split}
		&\mathcal{J}_R(x_i^L,x_i^{L^+},x_i^{L^-}) = \\
		&\max(0, m_t+\|b(x_i^L)-b(x_i^{L^+})\|_{\mathcal{H}}-\|b(x_i^L)-b(x_i^{L^-})\|_{\mathcal{H}})
	\end{split}
\end{equation}
where the constant parameter $m_t$ defines the margin difference metric between the relative similarity of the two pairs $(x_i^L,x_i^{L^+})$ and $(x_i^L,x_i^{L^-})$. That is to say, we expect the distance of the dissimilar pair $(x_i^L,x_i^{L^-})$ to be larger than the  distance of the similar pair $(x_i^L,x_i^{L^+})$ by at least $m_t$. $\|\cdot\|_{\mathcal{H}}$ denotes the weighted Hamming distance defined as:
\begin{equation}
	\mathcal{H}(x_i,x_j)=\sum_kW(C_{x_i},k)^2(b_k(x_i)\oplus b_k(x_j))
\end{equation}
where $C_{x_i}$ is the class index of image $x_i$. Note that in the weighted triplet ranking loss, the weights of anchor point $x_i^L$ are used to calculate the weighted Hamming distance. Because anchor point acts like query in the retrieval process, we treat anchor point's weights more importantly. Minimizing $\mathcal{J}_R$ can reach our goal to preserve the semantic ranking constraints provided by labels.

In equation~(\ref{rankingterm1}), the binary hash code and Hamming distance make it hard to directly optimize. Similar to NINH~\cite{7298947}, binary hash code $b(x)$ is relaxed with continuous real value hash code $h(x)$. Hamming distance is replaced by weighted Euclidean distance $\|\cdot\|_w$ defined as:
\begin{equation}
	\mathcal{W}(x_i,x_j)=\sum_kW(C_{x_i},k)^2(h_k(x_i)-h_k(x_j))^2
\end{equation}
Then equation~(\ref{rankingterm1}) can be rewritten as:
\begin{equation}
	\label{rankingterm2}
	\begin{split}
		&\mathcal{J}_R(x_i^L,x_i^{L^+},x_i^{L^-}) = \\
		&\max(0, m_t+\|h(x_i^L)-h(x_i^{L^+})\|_w^2-\|h(x_i^L)-h(x_i^{L^-})\|_w^2)
	\end{split}
\end{equation}
\subsubsection{The Classification Stream}
Besides the hash stream, we also design a classification stream connected with the representation layers. On one hand, jointly training the hash stream and the classification stream can improve the retrieval accuracy, which has been shown in previous work~\cite{xia2014supervised}. On the other hand, the trained classification stream can be used to generate the query-adaptive hash code weights, which will be introduced in next part. In the classification stream, a fully connected layer with $c$ neural nodes is connected with the representation learning layers, which predicts the probability of each class. Then softmax loss is used to train the classification stream:
\begin{equation}
	\label{classficicationError}
	\mathcal{J}_C(\theta)=-\dfrac{1}{m}[\sum_{i=1}^{m}\sum_{j=1}^{c}{I(y_j^i=1)\dfrac{e^{\theta_{j}x_i}}{\sum_{l=1}^{c}e^{\theta_{l}x_i}}}]
\end{equation}
where $\theta$ are parameters of the network, $m$ is the number of images in one batch, and $y_j^i\in\{0,1\},j=1,\cdots,c$ denotes whether image $x_i$ belongs to class $j$. Note that this is not a standard softmax loss, but a multilabel softmax loss, which can handle images with multiple labels. When only one element of $y_j^i$ is $1$, the above equation is equal to standard softmax loss. Incorporating the hash stream and the classification stream, the network can not only preserve the ranking information and semantic information, but also learn the bit-wise hash weights for different image classes.
\subsubsection{Network Training}
Forward and backward propagation schemes are used in the training phase. For the two streams in the network, we use a co-training method to tune the network parameters jointly. More specifically, in the forward propagation stage, the ranking error in the hash stream is measured by equation~(\ref{rankingterm2}), and the classification error in the classification stream is measured by equation~(\ref{classficicationError}). Then in the backward propagation stage, the network parameters are tuned by the gradient of each loss function. For the weighted triplet based ranking loss, the gradient with respect to $h(x_i^L)$, $h(x_i^{L^+})$ and $h(x_i^{L^-})$ are computed as:
\begin{equation}
	\label{gradient1}
	\begin{split}
		&\frac{\partial{\mathcal{J}_R}}{\partial{h(x_i^L)}} = 2W(C_x,:)^2\times(h(x_i^{L^-})-h(x_i^{L^+}))\times I_{c} \\
		&\frac{\partial{\mathcal{J}_R}}{\partial{h(x_i^{L^+})}} = 2W(C_x,:)^2\times(h(x_i^{L^+})-h(x_i^L))\times I_{c} \\
		&\frac{\partial{\mathcal{J}_R}}{\partial{h(x_i^{L^-})}} = 2W(C_x,:)^2\times(h(x_i^L)-h(x_i^{L^-}))\times I_{c} \\
		& I_{c} = I_{m_t+\|h(x_i^L)-h(x_i^{L^+})\|^2-\|h(x_i^L)-h(x_i^{L^-})\|^2>0}
	\end{split}
\end{equation}
Where $I_{c}$ is an indicator function, $I_{c} = 1$ if $c$ is true, otherwise $I_{c} = 0$. Then the gradient of each image is fed into the network to update parameters of each layer, including the hash layer and the weight layer.

For the softmax loss, the gradient with respect to $\theta_j$ is calculated as:
\begin{equation}
	\label{gradient2}
	\frac{\partial{\mathcal{J}_C}}{\partial{\theta_j}} = -\dfrac{1}{m}\sum_{i=1}^{m}[x^i(I(y_j^i=1)-\dfrac{e^{\theta_{j}x_i}}{\sum_{l=1}^{c}e^{\theta_{l}x_i}})]
\end{equation}
By equations~(\ref{gradient1}) and (\ref{gradient2}), these derivative values can be fed into the network via the back-propagation algorithm to update the parameters of each layer in the deep network. The training procedure is ended until the loss converges or a predefined maximal iteration number is reached. We briefly summarize the training process in Algorithm~\ref{trainalgo}. Note that after the network is trained, we can not only get the hash mapping functions, but also the hash weights associated with each bit.
\begin{algorithm}
\caption{Deep Hash Network Training Algorithm}
\label{trainalgo}
\begin{algorithmic}[1]
	\REQUIRE Image triplet set $\mathcal{T}$, parameter $m_t$
	\STATE Initialize weight layer with all $1$
	\STATE Randomly split $T$ into mini-batch
	\REPEAT
	\STATE Calculate outputs $h(x_i)$ for image $x_i$ by forward propagation
	\STATE Calculate triplet loss $J_R$ by equation~(\ref{rankingterm2})
	\STATE Calculate softmax loss $J_C$ by equation~(\ref{classficicationError})
	\STATE Calculate gradient of triplet loss by equation~(\ref{gradient1})
	\STATE Calculate gradient of softmax loss by equation~(\ref{gradient2})
	\STATE Update network parameter $\Theta$ by back propagation
	\UNTIL $J_R$ and $J_\theta$ converges
	\ENSURE Network parameters $\Theta$
\end{algorithmic}
\end{algorithm}
\subsection{Query-adaptive Image Retrieval}
\begin{algorithm}
	\caption{Query-adaptive Image Retrieval Algorithm}
	\label{retrievalalgo}
	\begin{algorithmic}[1]
		\REQUIRE Network parameters $\Theta$, query image $x_q$, database hash codes $B$
		\STATE Extracts weights parameters $W$ from $\Theta$
		\STATE Predict class probability $p(x_q)$
		\STATE Generate hash codes $b(x_q)$ by equation~(\ref{binarycode})
		\STATE Generate query-adaptive weights $w_q$ by equation~(\ref{queryweights})
		\STATE　Calculate weighted Hamming distance by equation~(\ref{weightham})
		\STATE Sort database images by weighted Hamming distance 
		\ENSURE Sorted image list
	\end{algorithmic}
\end{algorithm}
In the query stage, existing deep hashing methods~\cite{xia2014supervised,7298947,zhang2015bit} treat each hash bit equally, and they usually first map query image to binary hash codes and retrieve the images in the database by Hamming distance. However, Hamming distances are discrete values, which can not perform fine-grained ranking since a large amount of images may share the same distance to a query image. To address this issue, we propose the query-adaptive image retrieval approach. For a given query image $x_q$, we first generate real valued hash codes $h(x_q)$ by the output of hash layer, then the binary codes $b(x_q)$ are generated by equation~(\ref{binarycode}).

In order to perform query-adaptive fine-grained ranking, besides the hash codes, we also generate query-adaptive hash weights efficiently. Based on the trained network, we already obtain class-wise hash bit weights $W=\{w_{ij}| i = 1,\cdots,c, j = 1,\cdots,q\}$ for different image classes. The query-adaptive weights are generated rapidly as:
\begin{equation}
	\label{queryweights}
	w_q = W^Tp(x_q)
\end{equation}
where $p(x_q) = \{p_q^1,p_q^2,\cdots\,p_q^c\}$ is the predicted probability generated by the classification stream, in which $p_q^i$ indicates the probability that $x_q$ belongs to image class $i$. Equation~\ref{queryweights} means that we fuse the class-wise weighs by the probability of query $x_q$ belongs to each class, thus the generated hash bit weights can reflect the semantic property of the query image. Fine-grained image ranking can be performed by the weighted Hamming distance between the query and any image $x_i$ in the database:
\begin{equation}
	\label{weightham}
	dist(x_q,x_i) = \sum_{k=1}^{q}w_q(k)^2(b_k(x_q)\oplus b_k(x_i))
\end{equation}
where $q$ is the length of hash codes. We summarize the query-adaptive retrieval method in algorithm~\ref{retrievalalgo}. Note that the proposed query-adaptive image retrieval method is also very fast compared to original Hamming distance measurement. Equation~\ref{queryweights} is a simple matrix multiplication which is efficient. And in practice, the weighted Hamming distance only needs to be computed in a subset of hash codes, since we can firstly sort with the Hamming distance fast by $\oplus$ operation, and then compute the weighted distance in a subset within small Hamming distance (e.g. Hamming distance $\le2$). Thus the additional computation is very small compared to original Hamming distance ranking, and effective yet efficient fine-grained ranking can be performed.

\section{Experiments}
In this section, we will introduce our experiments conducted on four widely used datasets, which are CIFAR10, NUS-WIDE, MIRFLICKR and ImageNet datasets. We compare with eight state-of-the-art methods in terms of retrieval accuracy and efficiency to verify the effectiveness of our QaDWH approach. In addition, we also conduct baseline experiments to verify the separate contribution of proposed deep weighted hashing and query-adaptive retrieval.
\subsection{Datasets and Experimental Settings}
We conduct experiments on four widely used image retrieval datasets. Each dataset is split into query, database and training set,  we summarize the split of each dataset in Table~\ref{Traditionalsetting}, and detailed settings are as follows:
\begin{itemize}
	\item \textbf{CIFAR10} dataset consists of 60,000 $32\times 32$ color images from 10 classes, each of which has 6,000 images. Following~\cite{7298947,xia2014supervised}, 1,000 images are randomly selected as the query set (100 images per class). For the unsupervised methods, all the rest images are used as the training set. For the supervised methods, 5,000 images (500 images per class) are further randomly selected from the rest of images to form the training set.
	\item \textbf{NUS-WIDE}~\cite{chua2009nus} dataset contains nearly 270,000 images, each image is associated with one or multiple labels from 81 semantic concepts. Following \cite{7298947,xia2014supervised}, only the 21 most frequent concepts are used, where each concept has at least 5,000 images, resulting in a total of 166,047 images. 2,100 images are randomly selected as the query set (100 images per concept). For the unsupervised methods, all the rest images are used as the training set. For the supervised methods, 500 images from each of the 21 concepts are randomly selected to form the training set of total 10,500 images.
	\item \textbf{MIRFLICKR}~\cite{huiskes2008mir} dataset consists of 25,000 images collected from Flickr, and each image is associated with one or multiple labels of 38 semantic concepts. 1,000 images are randomly selected as the query set. For the unsupervised methods, all the rest images are used as the training set. For the supervised methods, 5,000 images are randomly selected from the rest of images to form the training set.
	\item \textbf{ImageNet}~\cite{ILSVRC15} dataset contains 1000 categories with 1.2 million images. ImageNet is a large dataset that can comprehensively evaluate the proposed approach and compared methods. Since the testing set of ImageNet is not publicly available, following~\cite{7298947}, we use the provided training set as the retrieval database, and the validation set as query set. For the training set of each hashing methods, we further randomly sampling 50,000 images from retrieval database as the training set (50 image per class).
\end{itemize}

\begin{table}[htb]
	\centering
		\caption{Split of each dataset}
		\label{Traditionalsetting}
		\begin{tabular}{c|c|c|c|c}
			\hline
			& CIFAR10 & NUS-WIDE & MIRFLICKR & ImageNet  \\ \hline
			Query    & 1,000 & 2,100    & 1,000    & 50,000         \\ \hline
			Database & 54,000 & 153,447   & 19,000    & 1,231,167        \\ \hline
			Training & 5,000 & 10,500   & 5,000    & 50,000         \\ \hline
		\end{tabular}
\end{table}
\begin{figure*}[tbh]
	\centering
	\includegraphics[width=\textwidth]{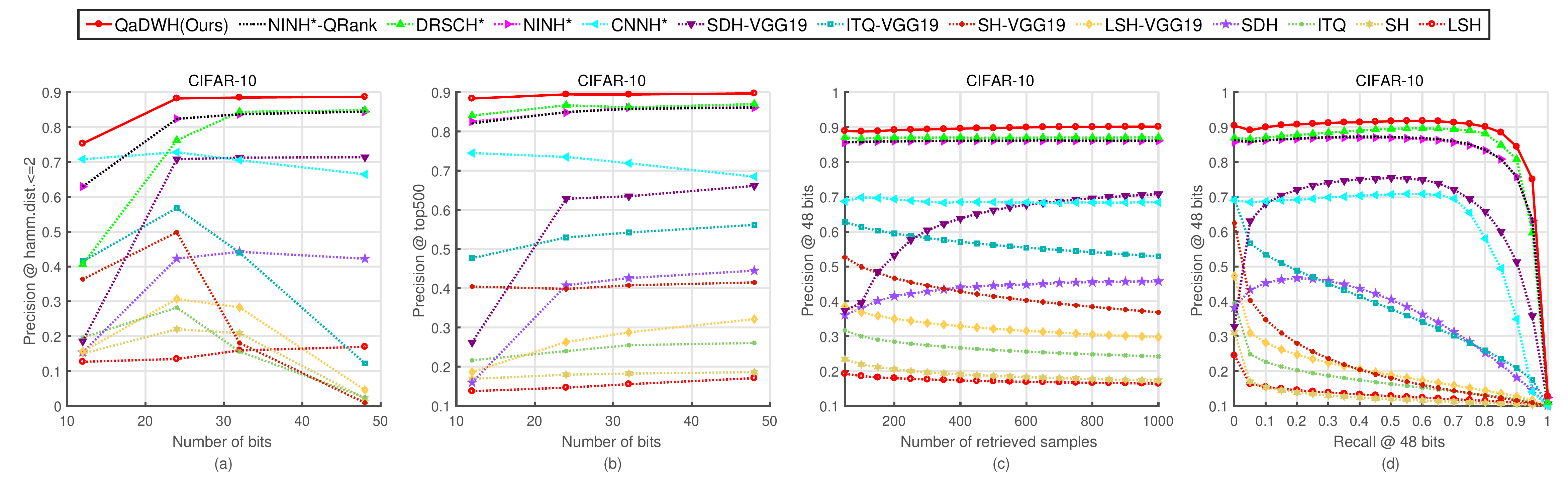}
	\caption{The comparison results on \textbf{CIFAR10}. (a) Precision within Hamming radius 2 using hash lookup; (b) Precision curves within top 500 retrieved samples w.r.t. different length of hash codes; (c) Precision curves with 48bit hash codes w.r.t. different number of top retrieved samples; (d) Precision-Recall curves of Hamming Ranking with 48bit.} 
	\label{CIFAR10Result}
\end{figure*}
\begin{figure*}[tbh]
	\centering
	\includegraphics[width=\textwidth]{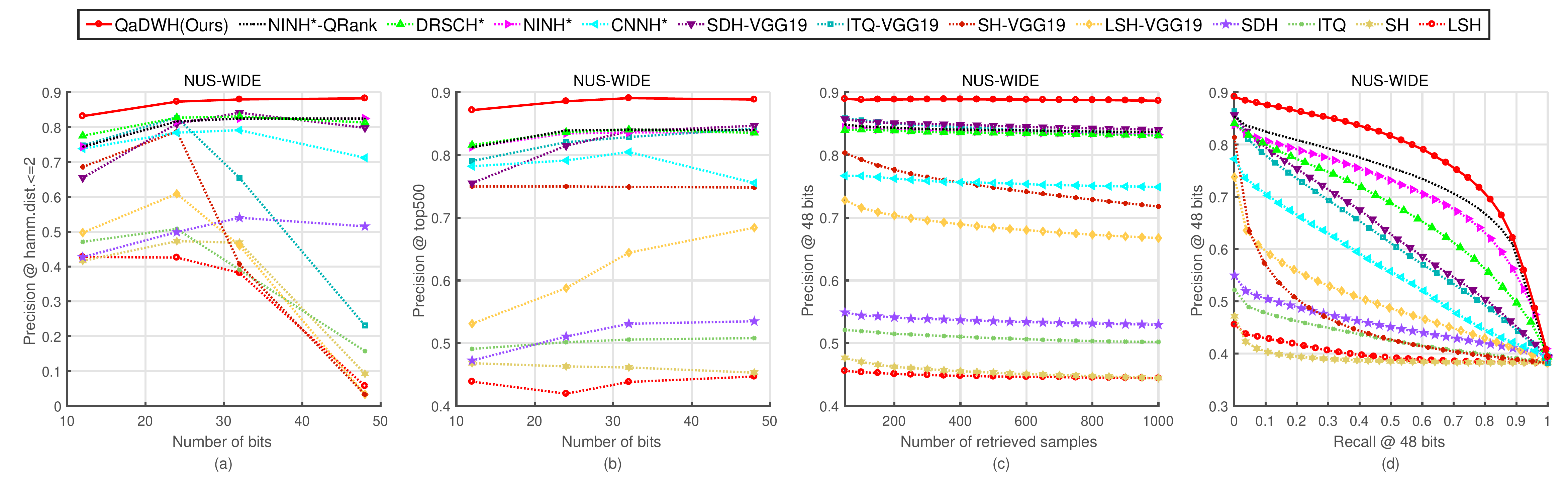}
	\caption{The comparison results on \textbf{NUS-WIDE}. (a) Precision within Hamming radius 2 using hash lookup; (b) Precision curves within top 500 retrieved samples w.r.t. different length of hash codes; (c) Precision curves with 48bit hash codes w.r.t. different number of top retrieved samples; (d) Precision-Recall curves of Hamming Ranking with 48bit.}
	\label{NUSWIDEResults}
\end{figure*}
\begin{figure*}[tbh]
	\centering
	\includegraphics[width=\textwidth]{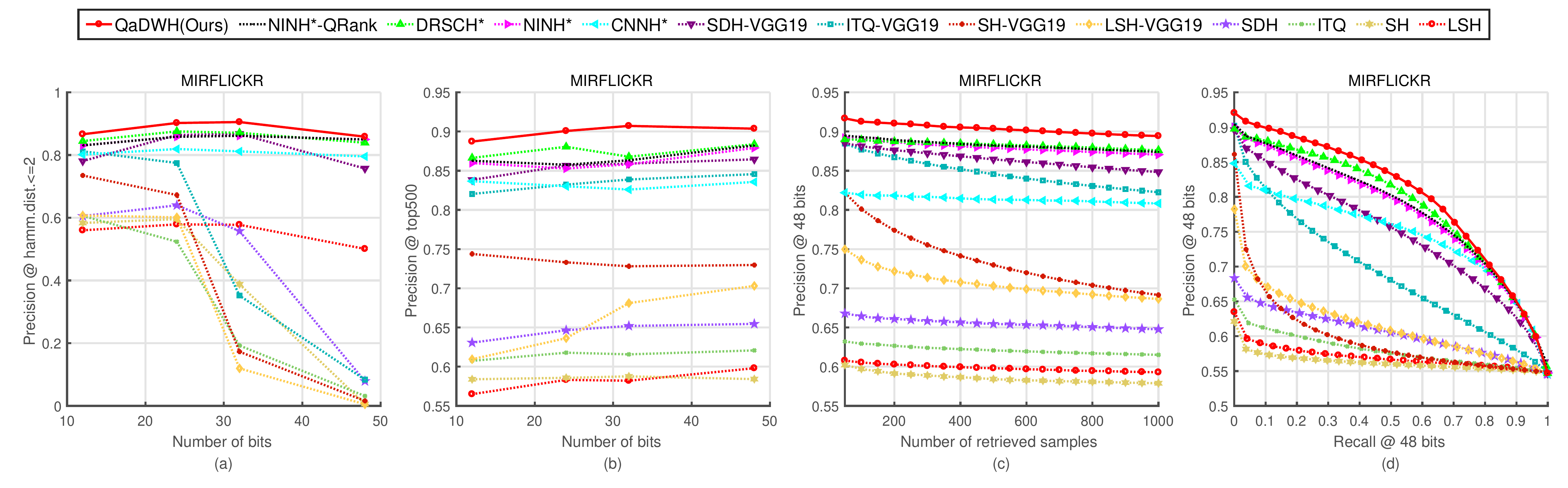}
	\caption{The comparison results on \textbf{MIRFLICKR}. (a) Precision within Hamming radius 2 using hash lookup; (b) Precision curves within top 500 retrieved samples w.r.t. different length of hash codes; (c) Precision curves with 48bit hash codes w.r.t. different number of top retrieved samples; (d) Precision-Recall curves of Hamming Ranking with 48bit.}
	\label{MIRFLCIKRresults}
\end{figure*}
\begin{figure*}[tbh]
	\centering
	\includegraphics[width=\textwidth]{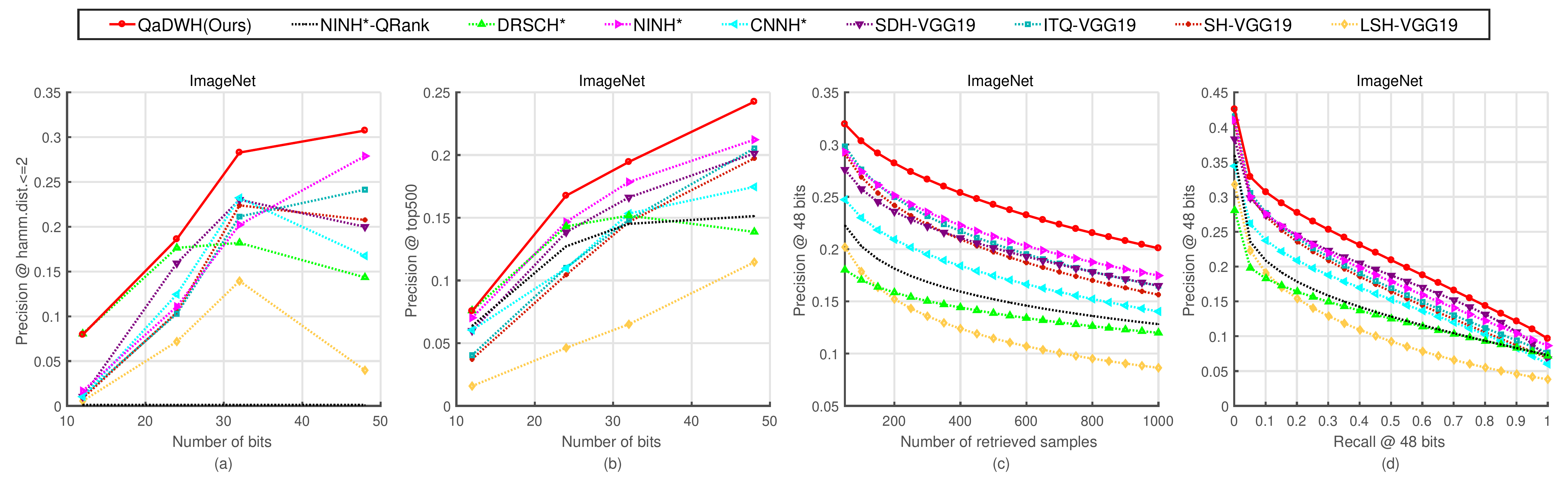}
	\caption{The comparison results on \textbf{ImageNet}. (a) Precision within Hamming radius 2 using hash lookup; (b) Precision curves within top 500 retrieved samples w.r.t. different length of hash codes; (c) Precision curves with 48bit hash codes w.r.t. different number of top retrieved samples; (d) Precision-Recall curves of Hamming Ranking with 48bit.}
	\label{ImageNetRresults}
\end{figure*}

\subsection{Evaluation Metrics and Compared Methods}
To objectively and comprehensively evaluate the retrieval accuracy of the proposed approach and the compared methods, we use 5 evaluation metrics: Mean Average Precision (MAP), Precision-Recall curves, precision curves of top k retrieved samples, precision within top 500 retrieved samples and precision within Hamming radius 2. The definitions of these evaluation metrics are defined as follows:
\begin{itemize}
	\item Mean Average Precision (MAP): MAP presents an overall measurement of the retrieval performance. MAP for a set of queries is the mean of average precision (AP) for each query, where AP is defined as:
	\begin{equation}
	AP=\frac{1}{R}\sum_{k=1}^{n}\frac{k}{R_k}\times rel_k
	\end{equation}
	where \textit{n} is the size of database set, R is the number of relevant images with query in database set, $R_k$ is the number of relevant images in the top k returns, and $rel_k=1$ if the image ranked at k-th position is relevant and 0 otherwise.
	\item Precision-Recall curves: The precisions at certain level of recall, we calculate Precision-Recall curves of all returned results.
	\item Precision curves of top k retrieved samples: The average precision of top k returned images for each query.
	\item Precision within top 500 retrieved samples: The average precision of the top 500 returned image for each query.
	\item Precision within Hamming radius 2: Precision curve of returned images with the Hamming distance smaller than 2 using hash lookup.
\end{itemize}

We compare the proposed QaDWH approach with eight state-of-the-art methods, including unsupervised methods LSH, SH and ITQ, supervised methods SDH, CNNH, NINH and DRSCH, and traditional query-adaptive hashing method QRank. The brief introductions of these 8 methods are listed below: 
\begin{itemize}
\item \textbf{LSH}~\cite{gionis1999similarity} is a data independent unsupervised method, which uses randomly generated hash functions to map image features into binary codes.
\item \textbf{SH}~\cite{weiss2009spectral} is a data dependent unsupervised method, which learns hash functions by making hash codes balanced and uncorrelated.
\item \textbf{ITQ}~\cite{gong2011iterative} is also a data dependent unsupervised method, which learns hash functions by minimizing the quantization error of mapping data to the vertices of a binary hypercube.
\item \textbf{SDH}~\cite{7298598} is a supervised method, which leverages label information to obtain hash codes by integrating hash code generation and classifier training.
\item \textbf{CNNH}~\cite{xia2014supervised} is a two-stage deep hashing method, which learns hash codes for training images in first stage, and trains a deep hashing network in second stage.
\item \textbf{NINH}~\cite{7298947} is a one-stage deep hashing method, which learns deep hashing network by a triplet loss function to measure the ranking information provided by labels.
\item \textbf{DRSCH}~\cite{zhang2015bit} is also a triplet loss based deep hashing method, which can further leverage hash code length by weighing each bit of hash codes.
\item \textbf{QRank}~\cite{ji2014query} is a traditional query-adaptive hashing method, which learns query-adaptive hash weights by exploiting both the discriminative power of each hash function and their complement for nearest neighbor search. QRank is state-of-the-art query-adaptive hashing method, which outperforms other weighted hashing methods (e.g. QsRank~\cite{zhang2012qsrank}, WhRank~\cite{zhang2013binary}).
\end{itemize}
\begin{table}[htb]
	\centering
	\caption{MAP scores with different length of hash codes on CIFAR10 dataset.}
	\label{cifar10table}
	\begin{tabularx}{0.48\textwidth}{c|YYYY}
		\hline
		\multirow{2}{*}{Methods} & \multicolumn{4}{c}{CIFAR10 (MAP)}  \\ \cline{2-5} 
		& 12bit & 24bit & 32bit & 48bit \\ \hline
		\textbf{QaDWH (ours)}                			  & \textbf{0.868} & \textbf{0.883} &\textbf{ 0.884} & \textbf{0.884}  \\ 
		NINH$\ast$-QRank~\cite{ji2014query}			   & 0.800 & 0.822 & 0.835 & 0.832 \\ \hline
		NINH$\ast$~\cite{7298947}                      & 0.792 & 0.818 & 0.832 & 0.830 \\ 
		DRSCH$\ast$~\cite{zhang2015bit}                & 0.820 & 0.852 & 0.850 & 0.851 \\ 
		CNNH$\ast$~\cite{xia2014supervised}            & 0.683 & 0.692 & 0.667 & 0.623 \\ \hline
		SDH-VGG19                   				  & 0.430 & 0.652 & 0.653 & 0.665 \\ 
		ITQ-VGG19                   				  & 0.339 & 0.361 & 0.368 & 0.375 \\ 
		SH-VGG19                    				  & 0.244 & 0.213 & 0.213 & 0.209 \\ 
		LSH-VGG19                   				  & 0.133 & 0.171 & 0.178 & 0.198 \\ \hline
		SDH~\cite{7298598}         				  & 0.255 & 0.330 & 0.344 & 0.360 \\ 
		ITQ~\cite{gong2011iterative}     		  & 0.158 & 0.163 & 0.168 & 0.169 \\ 
		SH~\cite{weiss2009spectral}    			  & 0.124 & 0.125 & 0.125 & 0.126 \\ 
		LSH~\cite{gionis1999similarity}  		  & 0.116 & 0.121 & 0.124 & 0.131 \\ \hline
	\end{tabularx}
\end{table}
\subsection{Implementation Details}
We implement the proposed approach based on the open-source framework Caffe~\cite{jia2014caffe}. The parameters of the first 18 layers in our network are initialized with the VGG-19 network~\cite{Simonyan14c}, which is pre-trained on the ImageNet dataset~\cite{ILSVRC15}. Similar initialization strategy has been used in other deep hashing methods~\cite{yaodeep,zhu2016deep}. For the weight layer, we initialize the weights with all $1$, because we treat each bit equally in the beginning of training. In all experiments, our network is trained with the initial learning rate of 0.001, we decrease the learning rate by 10 every 20,000 steps. And the mini-batch size is 64, the weight decay parameter is 0.0005. For the only parameter in our proposed loss function, we set $m_t=1$ in all the experiments.

For the proposed QaDWH, and compared methods CNNH, NINH and DRSCH, we use raw image pixels as input. The implementations of CNNH and DRSCH are provided by their authors, while NINH is of our own implementation. Since the representation learning layers of CNNH, NINH and DRSCH are different from each other, for a fair comparison, we use the same VGG-19 network as the base structure for deep hashing methods. And the network parameters of all the deep hashing methods are all initialized with the same pre-trained VGG-19 model, thus we can perform fair comparison between them. The results of CNNH, NINH and DRSCH are referred as CNNH$\ast$, NINH$\ast$ and DRSCH$\ast$ respectively.

For the query-adaptive method QRank, which uses image features and hash codes generated by other hashing methods as input. In order to compare QRank with proposed QaDWH approach fairly, we use the hash codes and features generated by deep hashing method NINH as the input of QRank, thus we denote the result of QRank as NINH$\ast$-QRank. The implementation of QRank is provided by the author.

For other compared traditional methods without deep networks, we represent each image by hand-crafted features and deep features respectively. For hand-crafted features, we represent images in the CIFAR10 and MIRFLICKR by 512-dimensional GIST features, and images in the NUS-WIDE by 500-dimensional bag-of-words features. For a fair comparison between traditional methods and deep hashing methods, we also conduct experiments on the traditional methods with the features extracted from deep networks, where we extract 4096-dimensional deep feature for each image from the pre-trained VGG-19 network. We denote the results of traditional methods using deep features by LSH-VGG19, SH-VGG19, ITQ-VGG19 and SDH-VGG19. The results of SDH, SH, and ITQ are obtained from the implementations provided by their authors, while the results of LSH are from our own implementation.
\begin{table}[htb]
	\centering
	\caption{MAP scores with different length of hash codes on NUS-WIDE dataset.}
	\label{nuswidetable}
	\begin{tabularx}{0.48\textwidth}{c|YYYY}
		\hline
		\multirow{2}{*}{Methods} & \multicolumn{4}{c}{NUS-WIDE (MAP)}  \\ \cline{2-5} 
		& 12bit & 24bit & 32bit & 48bit \\ \hline
		\textbf{QaDWH (ours)}               	  & \textbf{0.867} & \textbf{0.879} & \textbf{0.884} & \textbf{0.882} \\
		NINH$\ast$-QRank~\cite{ji2014query}	   & 0.813 & 0.836 & 0.835 & 0.833 \\ \hline
		NINH$\ast$~\cite{7298947}              & 0.808 & 0.827 & 0.827 & 0.827 \\ 
		DRSCH$\ast$~\cite{zhang2015bit}        & 0.814 & 0.829 & 0.832 & 0.824 \\ 
		CNNH$\ast$~\cite{xia2014supervised}    & 0.768 & 0.784 & 0.790 & 0.740 \\ \hline
		SDH-VGG19                    	  & 0.730 & 0.797 & 0.819 & 0.830 \\ 
		ITQ-VGG19                    	  & 0.777 & 0.800 & 0.806 & 0.817 \\ 
		SH-VGG19                     	  & 0.712 & 0.697 & 0.689 & 0.682 \\ 
		LSH-VGG19                    	  & 0.518 & 0.567 & 0.618 & 0.651 \\ \hline
		SDH~\cite{7298598}                & 0.460 & 0.510 & 0.519 & 0.525 \\ 
		ITQ~\cite{gong2011iterative}      & 0.472 & 0.478 & 0.483 & 0.476 \\ 
		SH~\cite{weiss2009spectral}       & 0.452 & 0.445 & 0.443 & 0.437 \\ 
		LSH~\cite{gionis1999similarity}   & 0.436 & 0.414 & 0.432 & 0.442 \\ \hline
	\end{tabularx}
\end{table}
\subsection{Experiment Results and Analysis}
\subsubsection{Experiment results on CIFAR10 dataset}
Table~\ref{cifar10table} shows the MAP scores with different length of hash codes on CIFAR10 dataset. Overall, the proposed QaDWH achieves the highest average MAP of 0.880, and consistently outperforms state-of-the-art methods on all hash code lengths. More specifically, compared with the highest deep hashing methods DRSCH$\ast$, which achieves average MAP of 0.843, the proposed QaDWH has an absolute improvement of 0.037. Compared with the highest traditional methods using deep features SDH-VGG19, which achieves an average MAP of 0.600, the proposed method has an absolute improvement of 0.280. While the highest traditional methods using hand-crafted features SDH achieves average MAP of 0.322, the proposed approach has an improvement of 0.558. And compared with the traditional weighted hashing method QRank, which achieves an average MAP of 0.822, the proposed QaDWH has an absolute improvement of 0.058. It's because QaDWH benefits from the joint training of hash codes and corresponding class-wise weights, while QRank can only learn the weights but cannot give feedback for learning better hash codes.

Figure~\ref{CIFAR10Result}(a) shows the precisions within Hamming radius 2 using hash lookup. The precision of proposed QaDWH consistently outperforms state-of-the-art methods on all hash code length, because QaDWH benefits from the joint training scheme and can generate better hash codes. The precision within top 500 retrieved samples is shown in Figure~\ref{CIFAR10Result}(b), the proposed QaDWH still achieves the highest precision, which demonstrates the effectiveness of fine-grained ranking. Figure~\ref{CIFAR10Result}(c) shows the precision curves of different number of retrieved samples on 48bit hash code, and the proposed QaDWH achieves the highest accuracy. Figure~\ref{CIFAR10Result}(d) demonstrates the precision-recall curves using Hamming ranking with 48bit codes. QaDWH still achieves the best accuracy on all recall levels, which further shows the effectiveness of proposed approach.

\begin{table}[htb]
	\centering
	\caption{MAP scores with different length of hash codes on MIRFLICKR dataset.}
	\label{Flickrtable}
	\begin{tabularx}{0.48\textwidth}{c|YYYY}
		\hline
		\multirow{2}{*}{Methods} & \multicolumn{4}{c}{MIRFLICKR (MAP)} \\ \cline{2-5} 
		& 12bit  & 24bit & 32bit & 48bit \\ \hline
		\textbf{QaDWH (ours)}               			   		& \textbf{0.791}  & \textbf{0.804} & \textbf{0.805} & \textbf{0.802} \\ 
		NINH$\ast$-QRank~\cite{ji2014query}	   			& 0.777 & 0.761 & 0.765 & 0.781 \\ \hline
		NINH$\ast$~\cite{7298947}                       & 0.772  & 0.756 & 0.760 & 0.778 \\ 
		DRSCH$\ast$~\cite{zhang2015bit}                 & 0.780  & 0.789 & 0.774 & 0.788 \\ 
		CNNH$\ast$~\cite{xia2014supervised}             & 0.763  & 0.757 & 0.758 & 0.744 \\ \hline
		SDH-VGG19                    			   		& 0.732  & 0.739 & 0.737 & 0.747 \\ 
		ITQ-VGG19                    			   		& 0.686  & 0.685 & 0.687 & 0.689 \\ 
		SH-VGG19                     			  		& 0.618  & 0.604 & 0.598 & 0.595 \\ 
		LSH-VGG19                    			   		& 0.575  & 0.584 & 0.604 & 0.614 \\ \hline
		SDH~\cite{7298598}                  	   		& 0.595  & 0.601 & 0.608 & 0.605 \\ 
		ITQ~\cite{gong2011iterative}        	   		& 0.576  & 0.579 & 0.579 & 0.580 \\ 
		SH~\cite{weiss2009spectral}         	   		& 0.561  & 0.562 & 0.563 & 0.562 \\ 
		LSH~\cite{gionis1999similarity}            		& 0.557  & 0.564 & 0.562 & 0.569 \\ \hline
	\end{tabularx}
\end{table}
\begin{table}[htb]
	\centering
	\caption{MAP scores with different length of hash codes on ImageNet dataset.}
	\label{imagenettable}
	\begin{tabularx}{0.48\textwidth}{c|YYYY}
		\hline
		\multirow{2}{*}{Methods} & \multicolumn{4}{c}{ImageNet (MAP)} \\ \cline{2-5} 
		& 12bit   & 24bit   & 32bit  & 48bit  \\ \hline
		\textbf{QaDWH (Ours)}              						& \textbf{0.090}   & \textbf{0.212}   & \textbf{0.245}  & \textbf{0.298}  \\ 
		NINH$\ast$-QRank~\cite{ji2014query}              & 0.078   & 0.170   & 0.198  & 0.208  \\ \hline
		NINH$\ast$~\cite{7298947}  						 & 0.076   & 0.162   & 0.197  & 0.236  \\ 
		DRSCH$\ast$~\cite{zhang2015bit}                  & 0.064   & 0.175   & 0.188  & 0.176  \\ 
		CNNH$\ast$~\cite{xia2014supervised}              & 0.076   & 0.151   & 0.204  & 0.230  \\ \hline
		SDH-VGG19                 						 & 0.075   & 0.182   & 0.216  & 0.261  \\ 
		ITQ-VGG19                						 & 0.054   & 0.151   & 0.201  & 0.268  \\ 
		SH-VGG19                 						 & 0.052   & 0.147   & 0.201  & 0.263  \\ 
		LSH-VGG19                						 & 0.027   & 0.079   & 0.110  & 0.182  \\ \hline
	\end{tabularx}
\end{table}
\subsubsection{Experiment results on NUS-WIDE dataset}
Table~\ref{nuswidetable} shows the MAP scores with different length of hash codes on NUS-WIDE dataset. Following~\cite{7298947,xia2014supervised}, we calculate the MAP scores based on top 5000 returned images. Similar results on NUS-WIDE can be observed, the proposed QaDWH still achieves the best MAP scores (average 0.878). QaDWH achieves an absolute improvement of 0.053 on average MAP compared to the highest deep hashing methods DRSCH$\ast$ (average 0.825). Compared with the highest traditional method using deep features SDH-VGG19, which achieves an average MAP of 0.794, QaDWH has an absolute improvement of 0.084. It is also interesting to observe that with the deep features extracted from VGG-19 network, the traditional method SDH achieves comparable results with deep hashing methods. And compared with QRank (average 0.829), the proposed QaDWH still achieves an  absolute improvement of 0.049, which shows that the proposed QaDWH method has the advantage of joint training hash code and corresponding weights.

Figure~\ref{NUSWIDEResults} (a), (b), (c) and (d) demonstrate the retrieval accuracy on NUS-WIDE. Similarly, the proposed QaDWH achieves the best accuracy on the 4 evaluation metrics, due to the joint training scheme and the fine-grained ranking for different queries.
\begin{table}[htb]
	\centering
	\caption{Comparison of the average testing time (Millisecond per Image) on four benchmark dataset by fixing the hash code length 48.}
	\label{ComparisonTestingTime}
	\begin{tabularx}{0.48\textwidth}{c|YYYY}
		\hline
		Methods     		& CIFAR10 & NUS-WIDE & MIRFLICKR & ImageNet \\ \hline
		QaDWH (ours)  		& 9.29    & 10.48   & 9.22      &  24.77 	\\ 
		NINH$\ast$-QRank    & 65.03   & 66.24   & 64.88     &  115.27  \\ \hline
		DRSCH$\ast$         & 9.11    & 10.04   & 9.14      &  23.98  \\ 
		NINH$\ast$          & 8.83    & 9.40    & 9.04      &  15.78  \\ 
		CNNH$\ast$          & 8.93    & 9.53    & 9.04      &  15.94   \\ \hline
		SDH-VGG19     		& 9.07    & 9.45    & 8.92      &  15.17   \\ 
		ITQ-VGG19     		& 8.91    & 9.39    & 8.76      &  15.33   \\ 
		SH-VGG19      		& 8.95    & 9.41    & 8.78      &  15.10   \\ 
		LSH-VGG19     		& 8.90    & 9.39    & 8;.76     &  15.15   \\ \hline
	\end{tabularx}
\end{table}
\begin{figure*}[tbh]
	\centering
	\includegraphics[width=\textwidth]{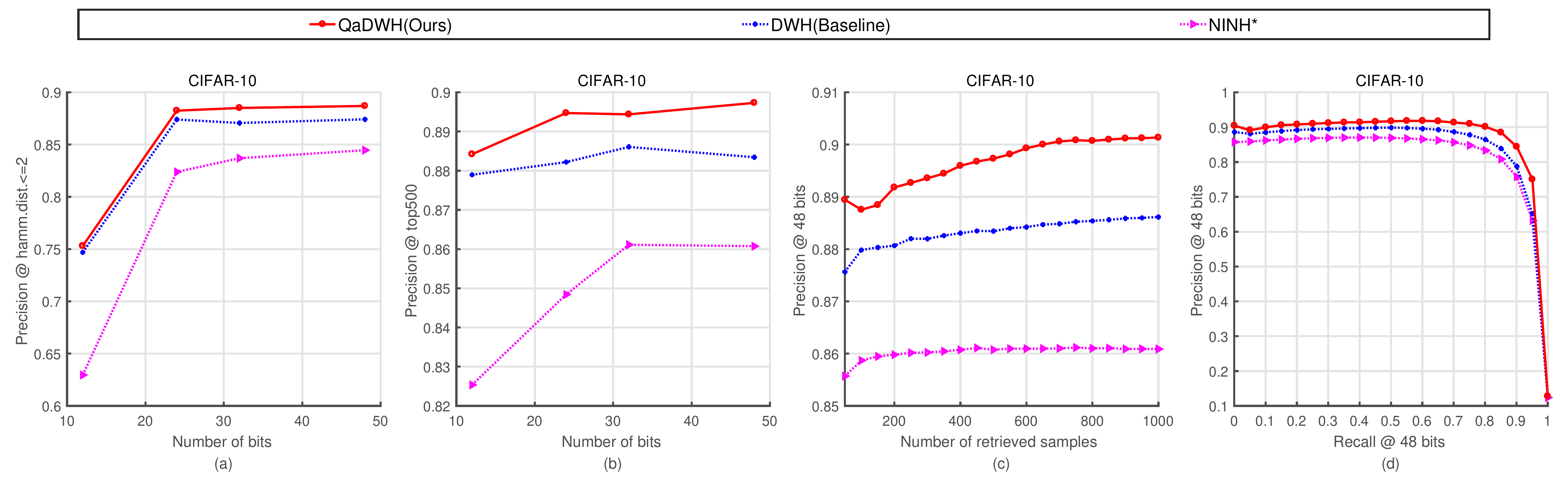}
	\caption{The comparison results of baseline methods on \textbf{CIFAR10}. (a) Precision within Hamming radius 2 using hash lookup; (b) Precision curves within top 500 retrieved samples w.r.t. different length of hash codes; (c) Precision curves with 48bit hash codes w.r.t. different number of top retrieved samples; (d) Precision-Recall curves of Hamming Ranking with 48bit.}
	\label{ciar10baseline}
\end{figure*}
\begin{figure*}[tbh]
	\centering
	\includegraphics[width=\textwidth]{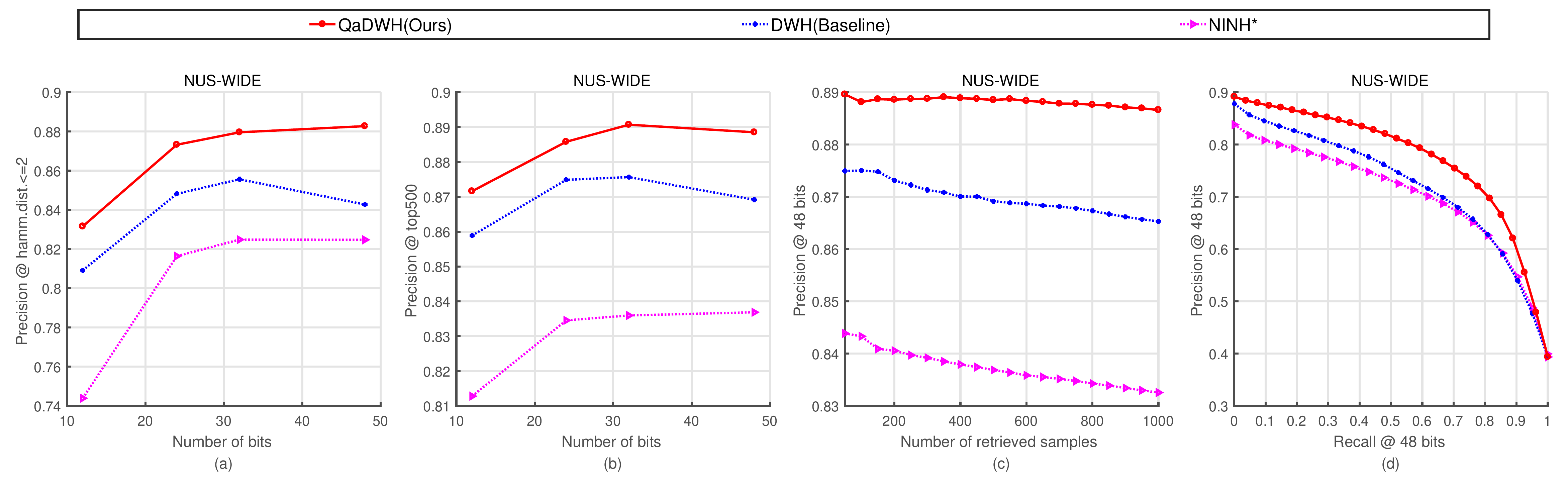}
	\caption{The comparison results of baseline methods on \textbf{NUS-WIDE}. (a) Precision within Hamming radius 2 using hash lookup; (b) Precision curves within top 500 retrieved samples w.r.t. different length of hash codes; (c) Precision curves with 48bit hash codes w.r.t. different number of top retrieved samples; (d) Precision-Recall curves of Hamming Ranking with 48bit.}
	\label{nuswidebaseline}
\end{figure*}
\begin{figure*}[tbh]
	\centering
	\includegraphics[width=\textwidth]{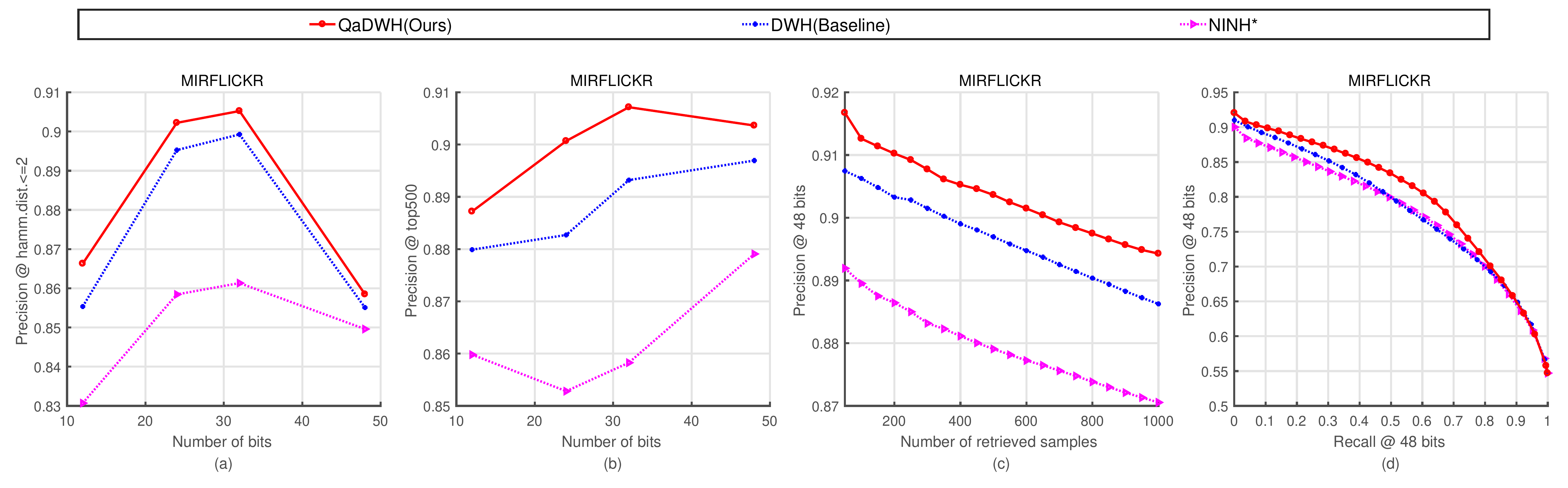}
	\caption{The comparison results of baseline methods on \textbf{MIRFLICKR}. (a) Precision within Hamming radius 2 using hash lookup; (b) Precision curves within top 500 retrieved samples w.r.t. different length of hash codes; (c) Precision curves with 48bit hash codes w.r.t. different number of top retrieved samples; (d) Precision-Recall curves of Hamming Ranking with 48bit.}
	\label{flickr25kbaseline}
\end{figure*}
\begin{figure*}[tbh]
	\centering
	\includegraphics[width=\textwidth]{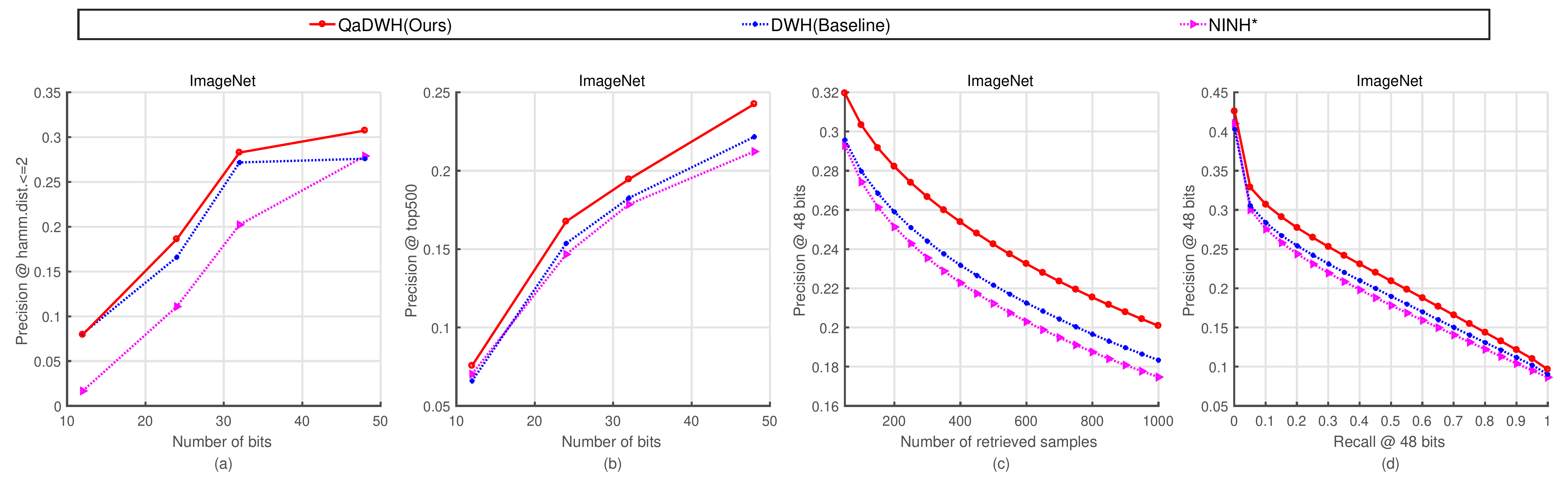}
	\caption{The comparison results of baseline methods on \textbf{ImageNet}. (a) Precision within Hamming radius 2 using hash lookup; (b) Precision curves within top 500 retrieved samples w.r.t. different length of hash codes; (c) Precision curves with 48bit hash codes w.r.t. different number of top retrieved samples; (d) Precision-Recall curves of Hamming Ranking with 48bit.}
	\label{imagenetbaseline}
\end{figure*}

\begin{table*}[htb]
	\centering
	\begin{scriptsize}
		\caption{MAP scores with different length of hash codes of baseline methods.}
		\label{baseline}
		\begin{tabularx}{\textwidth}{c|YYYY|YYYY|YYYY|YYYY}
			\hline
			\multirow{2}{*}{MAP} & \multicolumn{4}{c|}{CIFAR10}  & \multicolumn{4}{c|}{NUS-WIDE}  & \multicolumn{4}{c|}{MIRFLICKR} & \multicolumn{4}{c}{ImageNet} \\ \cline{2-17} 
			& 12bit & 24bit & 32bit & 48bit & 12bit & 24bit & 32bit & 48bit & 12bit  & 24bit & 32bit & 48bit & 12bit & 24bit & 32bit & 48bit \\ \hline
			QaDWH (ours)               & 0.868 & 0.883 & 0.884 & 0.884 & 0.867 & 0.879 & 0.884 & 0.882 & 0.791  & 0.804 & 0.805 & 0.802 & 0.090 & 0.212 & 0.245 & 0.298 \\ \hline
			DWH                  & 0.856 & 0.873 & 0.879 & 0.856 & 0.849 & 0.867 & 0.866 & 0.858 & 0.774  & 0.773 & 0.788 & 0.784 & 0.080 & 0.196 & 0.233 & 0.276 \\ \hline
			NINH$\ast$      	 & 0.792 & 0.818 & 0.832 & 0.830 & 0.808 & 0.827 & 0.827 & 0.827 & 0.772  & 0.756 & 0.760 & 0.778 & 0.076 & 0.162 & 0.197 & 0.236 \\ \hline
		\end{tabularx}
	\end{scriptsize}
\end{table*}

\subsubsection{Experiment results on MIRFLICKR dataset}
The MAP scores with different length of hash codes on MIRFLICKR dataset are shown in Table~\ref{Flickrtable}. The proposed QaDWH method achieves average MAP score of 0.800, which outperforms other deep hashing methods DRSCH$\ast$ (0.783), NINH$\ast$ (0.766) and CNNH$\ast$ (0.755). Compared with the highest traditional method using deep features SDH-VGG19, which achieves the average MAP of 0.739, QaDWH has an absolute improvement of 0.061. On MIRFLICKR, the proposed QaDWH method still outperforms traditional weighted hashing method QRank by 0.029, which shows the effectiveness of jointly training of hash codes and corresponding weights. Figure~\ref{MIRFLCIKRresults}(a) shows the precision within Hamming radius 2 using hash lookup, from which we can observe that the proposed QaDWH approach achieves the best result. Figure~\ref{MIRFLCIKRresults}(b) shows the precision curves within top 500 retrieved samples, and QaDWH achieves the highest precision due to the fine-grained retrieval. Figure~\ref{MIRFLCIKRresults}(c) and (d) demonstrate the top 1k results and Precision-Recall curve on 48bit hash code, the proposed QaDWH method still achieves the best results, which further shows the effectiveness of query-adaptive fine-grained ranking.

\subsubsection{Experiment results on ImageNet dataset}
The MAP scores with different length of hash codes on ImageNet dataset are shown in Table~\ref{imagenettable}, note that for this large scale dataset, we only report results of traditional methods using deep features. And for this large dataset, we calculate the MAP scores based on top 500 returned images due to the high computation cost of MAP evaluation. From Table~\ref{imagenettable} we can observe that the proposed QaDWH approach achieves best average MAP score of 0.211 on this challenging dataset. And compared with traditional weighted hashing method QRank, our proposed QaDWH achieves an absolute improvement of 0.048, and QRank cannot achieve stable improvements over NINH$\ast$ on this large dataset. Compared with the best deep hashing methods NINH$\ast$ on ImageNet dataset, the proposed QaDWH achieves an absolute improvement of 0.043. And on this large dataset, we can observe that traditional methods like SDH and ITQ achieve comparable results with deep hashing methods. Figure~\ref{ImageNetRresults} (a), (b), (c) and (d) demonstrate the retrieval accuracy on ImageNet. Similarly, the proposed QaDWH achieves the best accuracy on these four evaluation metrics, due to the joint training scheme of hash functions and corresponding weights and the fine-grained ranking for different query images.

\subsubsection{Comparison of Testing Time}
Besides the comparison of retrieval accuracy between different methods, we also compare the testing time of proposed approach and state-of-the-art methods. All the experiments are conducted on the same PC with NVIDIA Titan Black GPU, Intel Core i7-5930k 3.50GHz CPU and 64 GB memory. Typical retrieval process of hashing methods generally consists of three stages: Feature extraction, hash code generation and image retrieval among databases. We record time costs of each stage for different methods, the final testing time cost is the sum of three stages. Note that proposed QaDWH approach and other deep hashing methods are end-to-end frameworks, whose input are raw images and output are hash codes, while compared traditional hashing methods use image features as input. Thus for fair comparison, we use deep features for traditional methods. And for compared query-adaptive hashing method QRank, which uses image features and hash codes generated by other methods as input, its additional computation is query-adaptive weights calculation. The average testing time of different methods is shown in Table~\ref{ComparisonTestingTime}, we perform each hashing methods 5 times to calculate the average testing time. Comparing proposed QaDWH approach with other deep hashing methods, we can observe that QaDWH is a little slower but still comparable (less than 1 millisecond for small scale dataset, less than 10 milliseconds for large scale ImageNet dataset), which is expected since QaDWH uses relatively slower weighted Hamming distance. However, proposed QaDWH can achieve much better retrieval accuracy by a little time costs. Comparing proposed QaDWH with traditional query-adaptive method QRank, we can observe that proposed QaDWH is much faster than QRank, it's because QRank consumes much more time to calculate query-adaptive hash weights, while proposed QaDWH approach costs only a simple matrix multiplication to calculate query-adaptive weights. From the result table we can also observe that, the deep hashing methods and traditional hashing methods are comparable with each other in terms of testing time, since the time cost of hash code generation is only a matrix multiplication which is very fast, and all of them use Hamming distance that can be efficiently calculated by bit-wise XOR operation.

\subsection{Baseline Experiments and Analysis}
We also conduct two baseline experiments to further demonstrate the separate contributions of proposed deep weighted hashing and query-adaptive retrieval approach: (1) To verify the effectiveness of query-adaptive retrieval approach, we further perform experiments of using fixed weights by averaging learned class-wise weights, thus each query has the same hash code weights, we denote results of this baseline method as DWH. (2) To verify the effectiveness of deep weighted hashing, we further conduct experiments without using hash weights at all, which is equivalent to NINH method, we denote the results of NINH as NINH$\ast$. The MAP scores of baseline methods are shown in Table~\ref{baseline}. From the result table, we can observe that on all the four datasets, the DWH method outperforms NINH$\ast$, which shows that the learned class-wise weights can reflect the semantic property of different image classes, thus improve the retrieval accuracy. And QaDWH further outperforms DWH on all four datasets, which demonstrates that the query-adaptive image retrieval approach can further improve the retrieval accuracy. Figure~\ref{ciar10baseline} to \ref{imagenetbaseline} show other four evaluation metrics on CIFAR10, NUS-WIDE, MIRFLICKR and ImageNet datasets. From those figures we can clearly observe that DWH outperforms NINH$\ast$ and QaDWH outperforms DWH on those four evaluation metrics, which further demonstrate the effectiveness of proposed deep weighted hashing and query-adaptive retrieval approach.

\section{Conclusion}
In this paper, we have proposed a novel query-adaptive deep weighted hashing (QaDWH) approach. First, we design a new deep hashing network, which consists of two streams: the hash stream learns the compact hash codes and corresponding class-wise hash bit weights simultaneously, while the classification stream preserves the semantic information and improves hash performance. Second, we propose an effective yet efficient query-adaptive image retrieval approach, which first rapidly generates the query-adaptive hash weights based on the query's predicted semantic probability and class-wise weights, and then performs effective image retrieval by weighted Hamming distance. Experiment results show the effectiveness of QaDWH compared with eight state-of-the-art methods on four widely used datasets. In the future work, we intend to extend the deep weighted hashing scheme to a multi-table deep hashing framework, in which different weights are learned for different hash mapping functions.


%





\ifCLASSOPTIONcaptionsoff
  \newpage
\fi



\bibliographystyle{IEEEtran}
\bibliography{dwh}
\end{document}